\documentclass[journal,transmag]{IEEEtran}

\usepackage{amsmath,amsfonts}
\usepackage{mathtools,xparse}
\usepackage{bbm}
\usepackage{upgreek}
\usepackage{color}
\usepackage{algorithm}
\floatname{algorithm}{Alg.}     
\usepackage{algorithmic}    
\usepackage{subfig}
\usepackage{upgreek}

\title{Extracting 3D Vascular Structures from Microscopy Images using Convolutional Recurrent Networks}

\author{\IEEEauthorblockN{Russell Bates\IEEEauthorrefmark{1},
Benjamin Irving\IEEEauthorrefmark{1,4},
Bostjan Markelc\IEEEauthorrefmark{2},
Jakob Kaeppler\IEEEauthorrefmark{2},\\
Ruth Muschel\IEEEauthorrefmark{2},
Vicente Grau\IEEEauthorrefmark{1}, and
Julia~A.~Schnabel\IEEEauthorrefmark{1,3}
}
\IEEEauthorblockA{\IEEEauthorrefmark{1}Institute of Biomedical Engineering,\\Department of Engineering Science, University of Oxford, United Kingdom}
\IEEEauthorblockA{\IEEEauthorrefmark{2}CRUK/MRC Oxford Centre for Radiation Oncology, Department of Oncology, University of Oxford, United Kingdom}
\IEEEauthorblockA{\IEEEauthorrefmark{3}Division of Imaging Sciences and Biomedical Engineering, King’s College London, United Kingdom.}
\IEEEauthorblockA{\IEEEauthorrefmark{4}Perspectum Diagnostics, Oxford, United Kingdom}
}

\newcommand{\pder}[2][]{\frac{\partial#1}{\partial#2}}

\begin{document}
\maketitle

\begin{abstract}
\textbf{Abstract: }Vasculature is known to be of key biological significance, especially in the study of cancer. As such, considerable effort has been focused on the automated measurement and analysis of vasculature in medical and pre-clinical images. In tumors in particular, the vascular networks may be extremely irregular and the appearance of the individual vessels may not conform to classical descriptions of vascular appearance. Typically, vessels are extracted by either a segmentation and thinning pipeline, or by direct tracking. Neither of these methods are well suited to microscopy images of tumor vasculature. In order to address this we propose a method to directly extract a medial representation of the vessels using Convolutional Neural Networks. We then show that these two-dimensional centerlines can be meaningfully extended into 3D in anisotropic and complex microscopy images using the recently popularized Convolutional Long Short-Term Memory units (ConvLSTM). We demonstrate the effectiveness of this hybrid convolutional-recurrent architecture over both 2D and 3D convolutional comparators. 
\end{abstract}

\section{Introduction}
Advances in microscopy have led to an ever increasing ability to see into the minute detail of the tumor micro-environment. Using \textit{in vivo} fluorescence microscopy, it is possible to observe the development of capillary level vascularization of the tumor. This development of neovasculature via angiogenesis has been identified as one of the key processes in the development of a tumor \cite{Hanahan2011}. However, the closely packed and complex organization of this neovasculature makes automatic analysis extremely challenging. This is compounded by the fact that vascular networks are inherently 3-dimensional (3D) structures. Using multi-photon microscopy, it is possible to build up a 3D reconstruction of the vasculature. However, often, this leads to highly anisotropic voxel sizes and a drastically reduced ability to resolve detail in the \textit{z} dimension, compared to in the in-plane (\textit{x-y}) dimensions. Ultimately our aim is to be able to quantitatively track changes in the vasculature both temporally and between different tumors and different therapeutic conditions. In order to do this we require automated approaches for extracting information from the images. Typically, vascular imaging relies on the use of a contrast agent, which enhances the appearance of vasculature in the image. However, tumor vasculature is extremely leaky and irregularly perfused making this an unreliable approach. In addition to this, an important downstream application of the proposed method is to quantify levels of perfusion within the tumor. This requires a vessel extraction method that operates independently of levels of perfusion. However, imaging vasculature via the endothelium rather than the lumen means that our vasculature will follow a non-standard appearance model of a hollow tube rather than an enhanced cylinder. This renders a number of established vessel segmentation algorithms unusable for data of this kind. 

\begin{figure}[t]
\centering
\includegraphics[width=\linewidth]{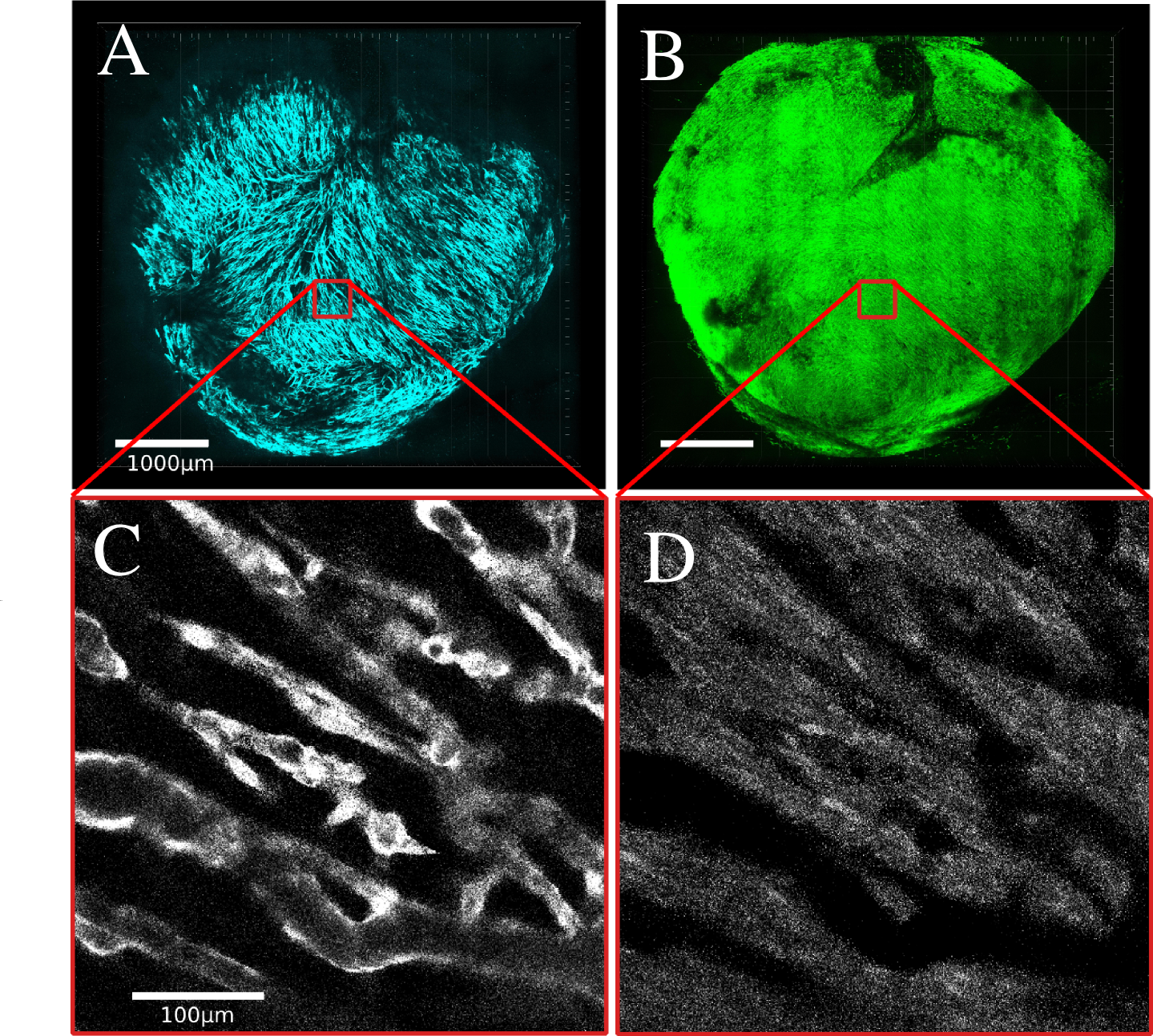}
\caption{A: Whole tumor, endothelium. B: Whole tumor, tumor cells. C: Single tile, endothelium. D: Single tile, tumor cells. Tile size is $512 \times 512 \times 70$ voxels, Approximately 200 tiles make up the entire image}
\end{figure}

The current state of the art in semantic image segmentation are the methods utilizing advances in deep learning, specifically deep Convolutional Neural Networks (CNNs). CNNs have proved extremely effective at a wide range of vision tasks due to their ability to build complex hierarchical representations of an image. By replacing the dense connectivity of the earlier neural networks with kernel convolutions, the number of model parameters is drastically reduced by enforcing a strong spatial prior. Repeated application of convolution and down-sampling allow the network to consider a large field-of-view, while only considering very local interactions at any given layer \cite{Simonyan2014,Krizhevsky2012}. Specific to image segmentation are the subclass of methods known as `fully convolutional' networks, which output an image of equal size to their input, transformed into some output domain \cite{Long2015}. These methods allow the reuse of many calculations by applying the same spatial priors applied to feature extraction to the labeling. Although CNNs have traditionally been applied to extremely large public databases such as for the ImageNet competition \cite{Russakovsky2015}, in the field of Medical Imaging there has been great interest in attempting to apply these same techniques to more bespoke applications with much smaller datasets \cite{Greenspan2016}.

Although CNNs have been extended to consider 3D interactions, we believe that this may be sub-optimal for applications such as ours, where the image does not represent an isotropic sampling of the domain. Since both the number of parameters and the computational complexity will increase exponentially with the effective dimension for convolutions, we restrict ourselves to convolution only in the information-rich \textit{x-y} plane. However, there is important contextual information contained in neighboring planes that we must exploit in order to produce a satisfactory extraction of the vasculature. In order to tap into this information without overly burdening ourselves with model complexity, we will use a variant of the Long Short-Term Memory (LSTM) recurrent neural network units \cite{Hochreiter1997}. The recently popularized Convolutional Long Short-Term Memory (ConvLSTM) units \cite{SHI2015} provide a way to model sequences of images. These networks modify the LSTM model of recurrent neural networks, replacing their inner product operations on feature vectors with kernel convolution operations on images. In this work we will apply these ConvLSTM units to tie together the detection results from each plane to create a meaningful 3D structure for the vasculature present in the image volume. 

A common pipeline for the extraction of vasculature is segmentation followed by a medial axis extraction, however, this is not always possible. In images containing tightly packed vascular structures it may be the case that a segmentation will be unable to separate extremely close, or even touching, sections of vasculature. If we then attempt to extract the vascular topology from this segmentation, it may not represent the true network structure of the underlying vasculature. In order to address this, we propose to learn to extract the medial representation of the vasculature directly, by training our network to reproduce skeletons of the vasculature, derived from manual segmentations, rather than training against the segmentations themselves. Difficulty arises from the fact that the medial representation of a 3D structure inherently requires a 3D view of the domain, and unlike segmentations it is extremely sparse. This cannot be achieved by just applying a CNN to each slice in our image. Instead, we propose to link together a shared CNN network applied to each image slice using ConvLSTM layers. This provides a mechanism for the sparse information to be shared between the feature extraction CNN networks on neighboring slices, even when no information about the skeleton may be present directly for that slice of the image. In addition to this we will analyze the role played by different loss functions and also the importance of applying our recurrent layers in a bidirectional sense, as has been explored in other applications such as video analysis \cite{GravesHybridLSTM}. 

\section{Related Work}

A number of authors have attempted to apply Deep Learning methods to segmentation of vasculature. However, the vast majority of effort in this area is applied to segmentation of vasculature in optical imaging of the retinal fundus  \cite{Liskowski2016,Fang2015,Li2016}. Prentašic \emph{et al. } have applied deep learning methods to coherence microscopy images of micro-vasculature \cite{Liskowski2016} but they consider only 2D images in which the vasculature already has an enhanced appearance. Teikari \emph{et al. } have applied CNNs to the task of segmentation in 3D fluorescence microscopy images of micro-vasculature, similar to ours \cite{Teikari2016}. However, they consider contrast enhanced vasculature, and only concern themselves with the segmentation of the vasculature, rather than extracting a medial representation. 3D CNNs have been applied successfully to brain lesion segmentation by Kamnitsas \emph{et al.}  \cite{Kamnitsas2016} with their Deep Medic architecture. 3D CNNs have also been explored in the V-Net architecture of Milletari \emph{at al.} \cite{Milletari2016} for prostate segmentation in MRI and the 3D extension of U-Net \cite{Cicek2016}.

ConvLSTMs were originally proposed by Shi \textit{et al.} \cite{SHI2015} for weather forecasting. 
They have recently been applied to a similar domain to ours in 3D microscopy by Chen \emph{et al.} \cite{Chen2016}. In this work they combine a stacked $k$U-net with a deep ConvLSTM architecture. However, they use this as a means to refine the segmentation, rather than fundamentally altering the output as we do. In addition to the convolutional LSTM, others have suggested the use of orthogonal LSTM units to traverse 3D image volumes, such as in the PyraMiD-LSTM work by Stollenga \emph{et al.} \cite{Stollenga2015}. Other convolutional-recurrent hybrids have also been explored in medical imaging, such as the work by Poudel \textit{et al.} \cite{Poudel2016}, who apply recurrence to the inner layers of a U-Net inspired architecture, for cardiac MRI segmentation.

The majority of classical vessel segmentation and extraction techniques rely on the tubular appearance of enhanced vasculature within an image. By considering the image pixel intensity as a hyper-surface N+1-dimensional space, we can consider a bright tubular structure as a ridge in this surface. This can be quantified by computing a local Hessian matrix for the surface, $\mathcal{H}_{ij} = \pder[I]{x_{ij}}$. Using considerations of the eigen-system of $\mathcal{H}$ it is possible to construct many vessel targeted segmentation methods. A thorough review of these methods may be found in the review papers by Lesage \cite{Lesage2009} and by Kirbas and Quek \cite{Kirbas2003}. However, as our vasculature does not present a regular tubular geometry or appearance as in most applications, we do not consider this to be a traditional vessel segmentation task. 

Skeletonization is usually applied as a post-processing step to binary segmentation images. The most common approaches consist of some variation of homotopic thinning \cite{Pudney1998}. A skeletal representation jointly describes the topology and the structure of our branching system of vasculature. This is a crucial stage in the process of quantitatively analyzing the vasculature as it allows us to attach rich, extracted features to some minimal representation of the vasculature.  A detailed review of the current state-of-the-art in 3D skeletonization algorithms has been provided by Tagliasacchi \emph{et al.} \cite{Tagliasacchi2016}. As they identify in their review, the skeleton of a volumetric structure is not a uniquely defined representation, and many such skeletons of a single volume may be equally valid. It is our view that, in light of this, it is beneficial to pass the decision making about what should be included in the vascular skeleton to the most `intelligent' part of our pipeline, namely the neural network.

In this work we will demonstrate the ability of hybrid convolutional-recurrent architectures to approach direct 'skeletonization', rather than relying heavily on thinning or tracking algorithms. We believe that by learning to extract centerlines directly in the machine learning task, we remove the degree of subjectivity which is involved in the skeletonization and pruning pipelines that are popular in many vessel extraction methods. 

\section{Materials}

In this section we will outline the data used for the experiments in this paper. We perform experiments on both real, pre-clinical data as well as synthetic data.

\textbf{Tumor Microscopy:} For our experiments on real data we have delineated vasculature from images acquired using high resolution fluorescence multi-photon microscopy. This approach achieves a theoretical lateral resolution of $0.4\upmu\mbox{m}$ and an axial resolution of $1.3\upmu\mbox{m}$. Voxels are sized $5\upmu\mbox{m}$ in the $z$ direction and $0.83 \upmu\mbox{m} \times 0.83 \upmu\mbox{m}$ in the $x$-$y$ plane. Imaging was performed using an abdominal window chamber model in mice. This allows for intra-vital imaging of the tumors \cite{Ritsma2013}. The abdominal window chamber was surgically implanted in transgenic mice on C57Bl/6 background that had expression of red fluorescent protein tdTomato in the endothelial cells only. The murine colon adenocarcinoma MC38 tumors with expression of green fluorescent protein (GFP) in the cytoplasm were induced by injecting $5\upmu\mbox{l}$ of dense cell suspension in a 50/50 mixture of saline and matrigel (Corning, NY, USA). The images of tumors were acquired 9 -– 14 days after tumor induction with Zeiss LSM 880 microscope (Carl Zeiss AG), connected to a Mai-Tai tunable laser (Newport Spectra Physics). We used an excitation wavelength of 940 nm and the emitted light was collected with Gallium Arsenide Phosphide (GaAsP) detectors through a 524 -- 546 nm bandpass filter for GFP and a 562.5 -- 587.5 nm bandpass filter for tdTomato and with a multi-alkali PMT detector through a 670 -- 760 bandpass filter for Qtracker® 705. A 20x water immersion objective with NA of 1.0 was used to acquire a Zstacks-TileScan with dimensions of $512 \times 512$ pixels in x and y, and approximately 70 planes in z, with a z step of $5\upmu\mbox{m}$. Each tumor is covered by approximately 100-200 tiles, depending on the size. All animal studies were performed in accordance with the Animals Scientific Procedures Act of 1986 (UK) and Committee on the Ethics of Animal Experiments of the University of Oxford.
The advantage of using both a labeled blood-pool based agent (Qtracker\textregistered 705) and transgenic mouse model with fluorescently labeled endothelium is that it allows us to assess the functional behaviour of the tumor vasculature. Skeletons are derived by first producing a manual segmentation of the vasculature on each slice. This is then followed by a centerline extraction of this manual segmentation in 3D, using the NeuTube software package \cite{Feng2015}. Extracted skeletons are then manually pruned and refined using the NeuTube software package. The training dataset consists of 25 manually segmented tiles of size $512 \times 512 \times 70$, which are then subdivided, with overlap, to form a larger training set of smaller image stacks. We hold back 10\% of our training data for validation and hyper-parameter tuning. The tiles were taken from 4 different tumors.

\begin{figure}
\centering
\includegraphics[width=\linewidth]{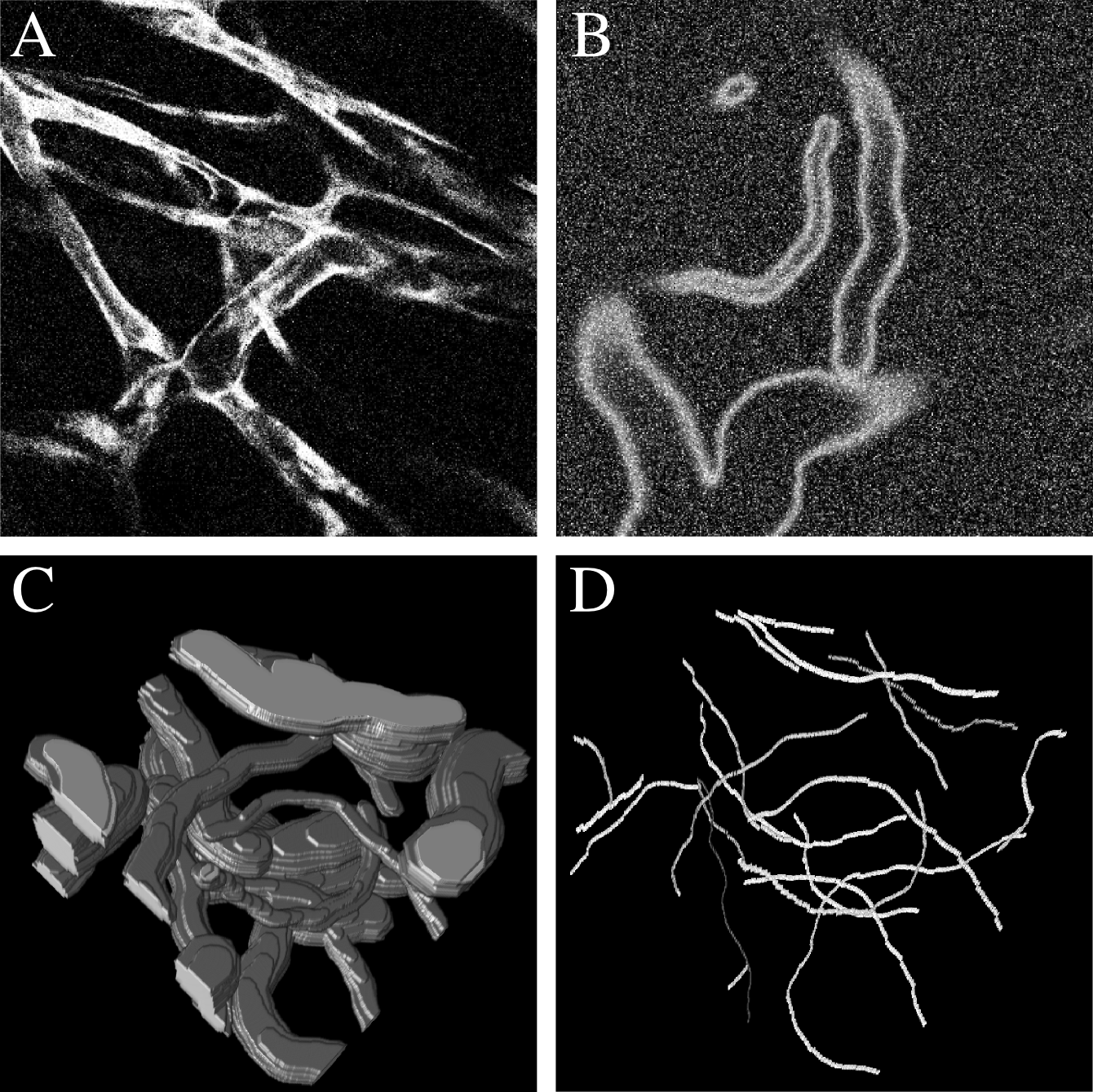}
\caption{(A) Real endothelium channel image, (B) Synthetic data image, (C) 3D rendering of synthetic segmentation, (D) 3D rendering of synthetic 3D skeleton.}
\label{fig:synthetic_data}
\end{figure}

\textbf{Synthetic data:} In order to test our method against a known ground truth, we have also generated a synthetic dataset that presents the same issues as our real data. The synthetic data consists of a number of hollow, tubular structures that are tightly packed and represented with anisotropic voxel sizes, comparable to the real data. The vessels are generated by first iteratively growing the centerlines throughout the volume. Then the segmentation mask is produced by dilating this skeleton to some randomly selected radius (chosen to reflect the range of sizes visible in the real data). We then generate two distance maps $d_1$ and $d_2$ that represent the distance from the nearest foreground and nearest background respectively. The endothelium is then generated from these distance maps:
\[
E = \exp(-d_1 / \sigma_1 ) \exp(-d_2 / \sigma_2),
\]
where $\sigma_1$ and $\sigma_2$ are tuned to give a qualitative appearance similar to our data. 
 Due to the anisotropy, many intersections between vessels are visible and the algorithm must reconstruct the 3D branching structure. We add Gaussian noise, whose variance is tuned to the background noise in the real images. We also add `Salt and Pepper' and Poisson noise to simulate the detector noise present in fluorescence microscopy. Slice `jitter' is also added to simulate the slight slice misalignment present in the real data, due to mouse breathing. Examples of this data are shown in Figure \ref{fig:synthetic_data}.

For the synthetic experiments we train on 20 image volumes, with size $512 \times 512 \times 40$, generated in this manner. Each volume is subdivided into tiles of size $128 \times 128 \times 16$. Testing was performed on 20 separate volumes, generated identically to the training volumes. 

\section{Methods}
Here we will give a description of the network architectures used, as well as details on training and the loss functions considered.

\subsection{Architectures}
\label{sec:architectures}

\textbf{CNN: } For the CNN portion of the network we use a U-Net style architecture \cite{Ronneberger2015} with 2 convolutional layers at each pooling level. Each convolutional layer is followed by a Batch Normalization \cite{Ioffe2015} layer and then a (Leaky) Rectified Linear Unit (LReLU) which is defined as follows.

\begin{equation}
LReLU(x; \alpha) = \left\{
\begin{array}{rl}
\alpha x & x < 0 \\
x & \mbox{otherwise}
\end{array} \right.
\end{equation}

The Leaky ReLU modification attempts to avoid the `dying ReLU' problem by maintaining a small slope in the negative portion whilst retaining the piecewise constant gradient, which provides the efficiency for ReLU activations. We use this as our highly background dominated target data can lead to many `dead' ReLU units, which we are unable to recover from during training.

\begin{figure}
\centering
\includegraphics[width=8cm]{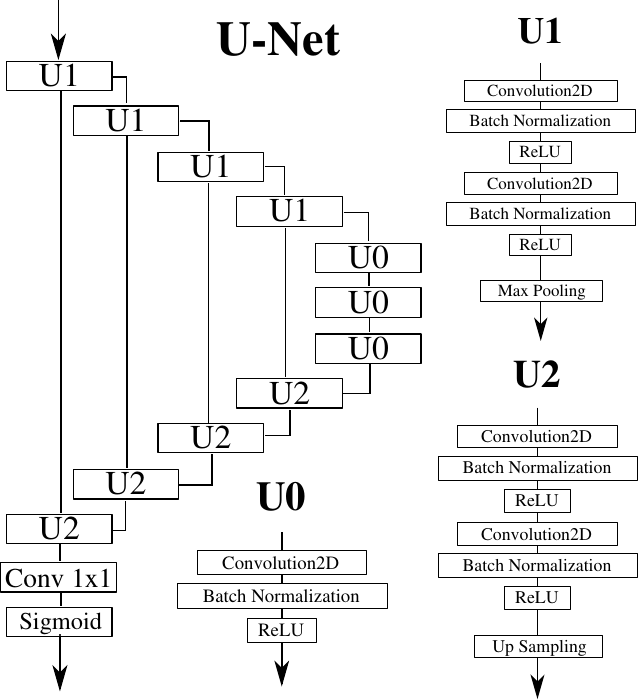}
\caption{Layer arrangement for a each depth layer of the U-Net architecture.}
\label{fig:unetfull}
\end{figure}

The full architecture is described in Figure \ref{fig:unetfull}. A sequence of down sampling units (\textbf{U1}) provide a broader and broader view on the image. In this coarse-grained view, a sequence of \textbf{U0} units extract rich features. These are then up-sampled by a sequence of \textbf{U2} units, whilst concatenating these new features with the features from the corresponding scaled \textbf{U1} unit. 

The final feature map is reduced using a $1\times1$ convolution before being passed through a sigmoidal activation:

\begin{equation}
\sigma(x) = \frac{1}{1 + e^{-x}}.
\end{equation}

This produces a feature map of the same spatial dimensions as the input image. 

All convolutions are performed using $3 \times 3$ kernels. Max pooling is performed over a $2 \times 2$ region and up-sampling is performed with stride $2$. Weights are initialized from a uniform distribution with bounds as suggested by Glorot \textit{et al.} \cite{Glorot}


\textbf{ConvLSTM:} The Convolutional LSTM modifies the traditional LSTM architecture \cite{Hochreiter1997} by replacing the inner product operations with kernel convolutions, therefore allowing us to efficiently operate on images, as in a CNN. The full state equations for this unit can be found in the work by Shi \emph{et al.} \cite{SHI2015}.


\begin{figure}[t]
\centering
\includegraphics[width=4cm]{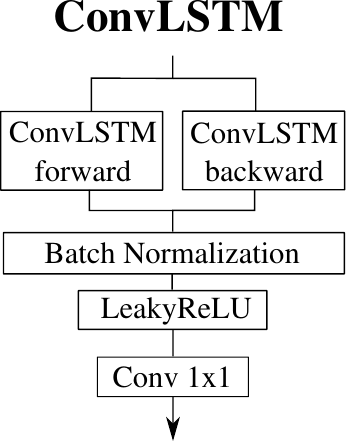}
\caption{Layer arrangement for a each depth layer of the U-Net architecture. }
\label{fig:convlstmunit}
\end{figure}

We apply each ConvLSTM bidirectionally, concatenating the outputs. Each ConvLSTM layer uses 20 units. Using the design principles of Network-in-Network \cite{Lin2013} to reduce dimensionality, we follow each bidirectional stack by a $1\times1$ convolution. For these experiments we explore this unit applied in both a `deep' and `shallow' configuration. 
\subsubsection{`Deep' configuration:}
\label{sec:deep_clstm}
For this architecture we arrange the ConvLSTM units (Figure \ref{fig:convlstmunit} in a $U$-shaped configuration as in the CNN network a modified U-Net configuration has also been employed by Chen \emph{et al.} \cite{Chen2016}. We stack two units, followed by a max pooling layer, followed by a further two units, followed by an up-sampling layer, followed by a final two units. The output from this is then passed through a $1\times1$ convolution to compress the features before a sigmoidal activation to provide the final output.
\subsubsection{`Shallow' configuration}
\label{sec:shallow_clstm}
For this configuration we use just a single ConvLSTM unit, as shown in Figure \ref{fig:convlstmunit}. 

\textbf{Combination: } In order to combine these networks, we apply a shared copy of the CNN network to each slice of the 3D image volume. This produces a 2D skeleton detection on each slice. This new volume is then processed in parallel by the stacked ConvLSTM layers to produce the new, context aware 3D skeleton detection. 

\begin{figure*}
\centering
\includegraphics[width=\textwidth]{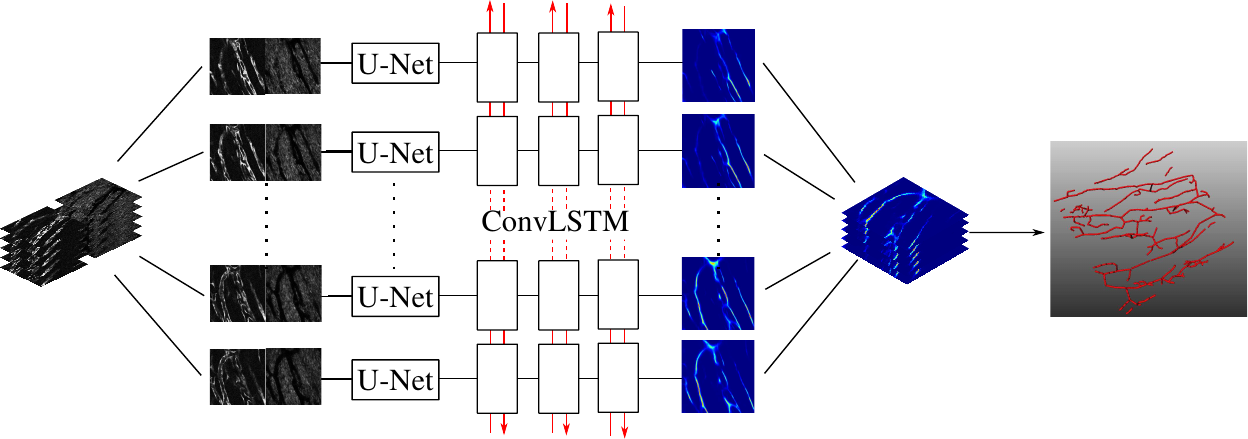}
\caption{Full combined architecture schematic for the U-Net2D + CLSTM networks. Image slices are passed independently through a shared CNN network, before being processed in parallel by stacked bidirectional ConvLSTM units. The final output will be approximately a 3D skeleton which can be refined to give our final skeletal representation. }
\label{fig:full_architecture}
\end{figure*}

\textbf{Training: } By design, it is possible that this network may be trained in an end-to-end fashion. However, depending on the size of the component architectures, it may be necessary to  separate the training for the two components of the network. Our training set consists of a number of manually annotated image volumes where all visible vasculature has been segmented. In order to train this end-to-end, we train against our extracted skeletons. 

In order to train the architecture in a split fashion, we suggest performing a skeletonization in 2D for each slice of the manual segmentations to train the CNN layers. The outputs from the trained CNN can then be used as input volumes for training the ConvLSTM layers, with the 3D thinned segmentations being the targets. Here we will test results training against a number of different loss functions.
A binary cross-entropy loss, given by:
\begin{equation}
\mathcal{L}_{bce}(x,y) = -\sum_i{y_i \log(x_i) + (1-y_i) \log(1-x_i)}.
\label{eq:bce}
\end{equation}
A Dice coefficient based loss:
\begin{equation}
\mathcal{L}_{d}(x,y) = 1 - 2\frac{x \cdot y + \delta}{|x| + |y| + \delta}
\label{eq:dice}
\end{equation}

An advantage of using a Dice coefficient based loss is that it implicitly focuses on the foreground class. This is advantageous for applications such as this where there exists a severe class imbalance between foreground (vessel centerline) and background. Here $\delta$ is a small positive value to keep the loss well behaved in the case that $|x| = |y| = 0$. For these experiments we set $\delta = 1$.

Another way to address this imbalance without discarding data is to use a class weighted loss function. Where foreground and background observations have their importance weighted according to the ratio of class imbalance. However, this has the effect of relatively under-weighting the background regions in the immediate vicinity of the thin foreground regions. We therefore propose a pixel-wise loss weighting, similar to Ronneberger \emph{et al.} \cite{Ronneberger2015}:

\begin{equation}
\mathcal{L}_{pbce}(x,y) = -\sum_i{W_i y_i \log(x_i) + W_i (1-y_i) \log(1-x_i)}. \\
\label{eq:wbce}
\end{equation}
\begin{equation}
W = (1-\beta) y \ast g_\sigma + \beta,
\end{equation}

where $g_\sigma$ represents a Gaussian kernel with standard deviation $\sigma$ and $\ast$ represents the kernel convolution operator. Here $\beta$ represents the ratio of class imbalance. 

In doing this we incorporate class imbalance while spreading its influence to the local vicinity of the thin foreground skeleton. In these experiments we trained our networks using the Adam optimizer \cite{Kingma2014} with a learning rate of $10^{-4}$. All training was performed using NVIDIA Tesla K40 12GB GPUs.

\textbf{Skeletonization: } A final stage in any pipeline for the analysis of vasculature may include a skeletonization process, to reduce any kind of detected representation to a topological skeletal representation for further analysis. Uncertainty and variability in our training set can result in skeletonization results that are thicker than a single pixel wide skeleton, we therefore account for this by performing a skeletonization as a post-processing. In these experiments we consider a simple homotopic thinning algorithm \cite{Pudney1998}, followed by a pruning phase to remove artifactual branches. A key advantage of our method is that although we still apply a thinning algorithm to finalize our results, the majority of `skeletonization' has already been performed by the neural network. 

\section{Results}

Here we will present the results of experiments both on synthetic data as well as real life microscopy imaging data. We will first briefly outline the architectures used for comparison. 

\textbf{U-2D: } A 2D U-Net architecture as described in Section \ref{sec:architectures} (CNN). In order to apply this to volumetric images, we apply the network to each slice in the $z$-stack independently and concatenate the results. Trainable weights: 200,000. 

\textbf{U-2D+CLSTM (S): } A shared 2D U-Net architecture is combined with a single (shallow) ConvLSTM unit with 32 filters as described in Section \ref{sec:shallow_clstm}. Trainable weights: 270,000.

\textbf{U-2D+CLSTM (D): } A shared 2D U-Net architecture is combined with a `$U$-shaped' (deep) ConvLSTM architecture as described in Section \ref{sec:deep_clstm}, with 32 filters per unit. Trainable weights: 390,000.

\textbf{U-3D: } A 3D U-Net architecture, similar to that described in Section \ref{sec:architectures} (CNN) but with all 2D convolution and pooling operations replaced with their 3D equivalent. Trainable weights: 580,000.

\textbf{CLSTM: } A `$U$-shaped' (deep) ConvLSTM architecture as described in Section \ref{sec:deep_clstm} (ConvLSTM). Trainable weights: 190,000.

We would like to explore the importance of the rich feature extraction of the CNN architectures (U-2D, U-3D) as well as the impact of the recursive architectures (CLSTM), both with and without a CNN component.

For each experiment we tune an optimal threshold on validation data for each architecture which is used to binarize the output from each network before skeletonization and analysis. 

\subsection{Results on synthetic data}

\subsubsection*{Experiment 1}
In order to provide a baseline for later comparisons we first compare the ability of our proposed network to a number of comparison networks to segment our synthetic data. We measure this using the Dice overlap, a commonly used segmentation evaluation metric, described in Equation \ref{eq:dice}. We compare a 2D U-Net, our U-Net + ConvLSTM, a ConvLSTM and a 3D U-Net network. All networks are trained to optimize a Dice coefficient based loss, given in Equation \ref{eq:dice}. For evaluation purposes we use the standard Dice coefficient, with $\beta = 0$. 

Results can be seen in Table \ref{tab:synth_seg_dice}. These networks are trained against the full segmentation masks. 

\begin{table}[h]
\centering
\begin{tabular}{l|c}
Architecture & Dice Score  \\
\hline
                U-2D & 0.73 \\
                U-2D+CLSTM (D)& \textbf{0.97} \\
                CLSTM & 0.87  \\
                U-3D &  0.95  \\
\end{tabular}
\caption{Results for Synthetic Experiment 1}
\label{tab:synth_seg_dice}
\end{table}

\begin{figure*}[t]
\includegraphics[width=\textwidth]{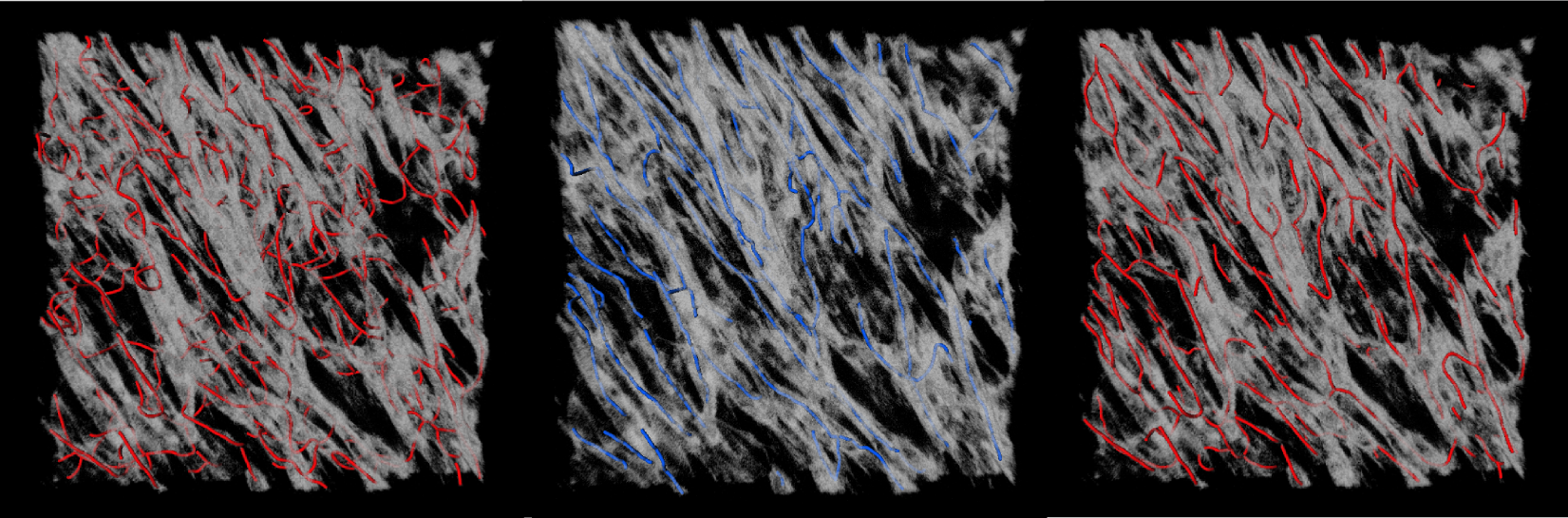}
\caption{Comparison of the U-Net2D method (left) with the proposed U-Net + ConvLSTM approach (right). A manually annotated ground truth is shown in blue (center). Detected vasculature is overlaid onto a volume rendering of the endothelium channel.}
\label{fig:results_compare}
\end{figure*}

\subsubsection*{Experiment 2}
Here we compare the ability of a number of networks to recreate the correct skeletal structure of the vessels in our synthetic data. In order to evaluate this we train all of our networks to recreate the 3D skeletons of the synthetic data, as can be seen in Figure \ref{fig:synthetic_data}. We train these against the weighted binary cross-entropy loss described in Equation \ref{eq:wbce} in order to account for the severe class imbalance. These networks are trained against the ground truth 3D skeletons. In order to evaluate skeleton distance we first threshold the output from each network and perform any necessary additional thinning until we have a single pixel-wide skeletal representation \cite{Pudney1998}. In order to compute the distance between skeletons we use the Modified Hausdorff Distance, proposed by Dubuisson \emph{et al.} \cite{Dubuisson1994}. 


We refer to this as `Skeleton Error' in the results section and can be interpreted as the average shortest distance between any point on the ground truth skeleton to some point on the target skeleton and vice versa. Here it is given in units of $\upmu$m. 

In addition to this we compute the `coverage' of the ground truth, the fraction of ground truth skeleton that is within some radius of the nearest bit of skeleton on the target to be compared. In this case we use a radius of 20$\upmu$m, as this is comparable in size to the radius of vasculature present in these images. 

The results of this experiment are shown in Table \ref{tab:synth_skel_haus}

\begin{table}[h]
\title{Skeleton Distance Comparison}
\centering
\begin{tabular}{l|c|c}
Architecture & Skeleton Error ($\upmu$m)  & Coverage \\ 
\hline
U-2D & 78.20 & 0.07 \\
U-2D+CLSTM (D) & \textbf{4.56}  & \textbf{0.96} \\
U-2D+CLSTM (S) & 7.42  & 0.88 \\
U-3D &  36.33  & 0.31 \\
CLSTM & 41.65  & 0.31 \\
\end{tabular}
\caption{Results for Synthetic Experiment 2}
\label{tab:synth_skel_haus}
\end{table}

\subsubsection*{Experiment 3}
Here we evaluate the importance of using a bidirectional ConvLSTM unit, rather than simply running in one direction. A single ConvLSTM unit would only be able to look through the image stack in a single direction, e.g. top-to-bottom or bottom-to-top. By applying ConvLSTM units in both directions and concatenating the outputs we gain the ability to consider changes occurring in both directions. The results of this experiment are shown in Table \ref{tab:bidirectional}

\begin{table}[h]
\title{Skeleton Distance Comparison}
\centering
\begin{tabular}{l|c|c}
Architecture & Skeleton Error ($\upmu$m)  & Coverage\\ 
\hline
Bidirectional & \textbf{4.56} & \textbf{0.96}  \\
Unidirectional & 6.47  & 0.93  \\
\end{tabular}
\caption{Results for Synthetic Experiment 3}
\label{tab:bidirectional}
\end{table}

\subsubsection*{Experiment 4}
Finally, we compare the effects of different loss functions for training to reproduce vessel skeletons. The loss functions we compare are the binary cross-entropy, the weighted binary cross-entropy and the Dice coefficient based loss. The results of this analysis can be seen in Table \ref{tab:loss_func}.
\begin{table}[h]
\title{Loss Function Comparison}
\centering
\begin{tabular}{l|c|c}
Architecture & Skeleton Error ($\upmu$m)  & Coverage \\ 
\hline
bce (Eq. \ref{eq:bce}) & 5.24 & 0.93 \\
w\_bce (Eq. \ref{eq:wbce}) & \textbf{4.56}  & \textbf{0.96} \\
Dice (Eq. \ref{eq:dice}) & n/a  & n/a \\
\end{tabular}
\caption{Results for Synthetic Experiment 4}
\label{tab:loss_func}
\end{table}

\subsection{Results on real microscopy data}

\begin{figure}
\centering
\includegraphics[width=\linewidth]{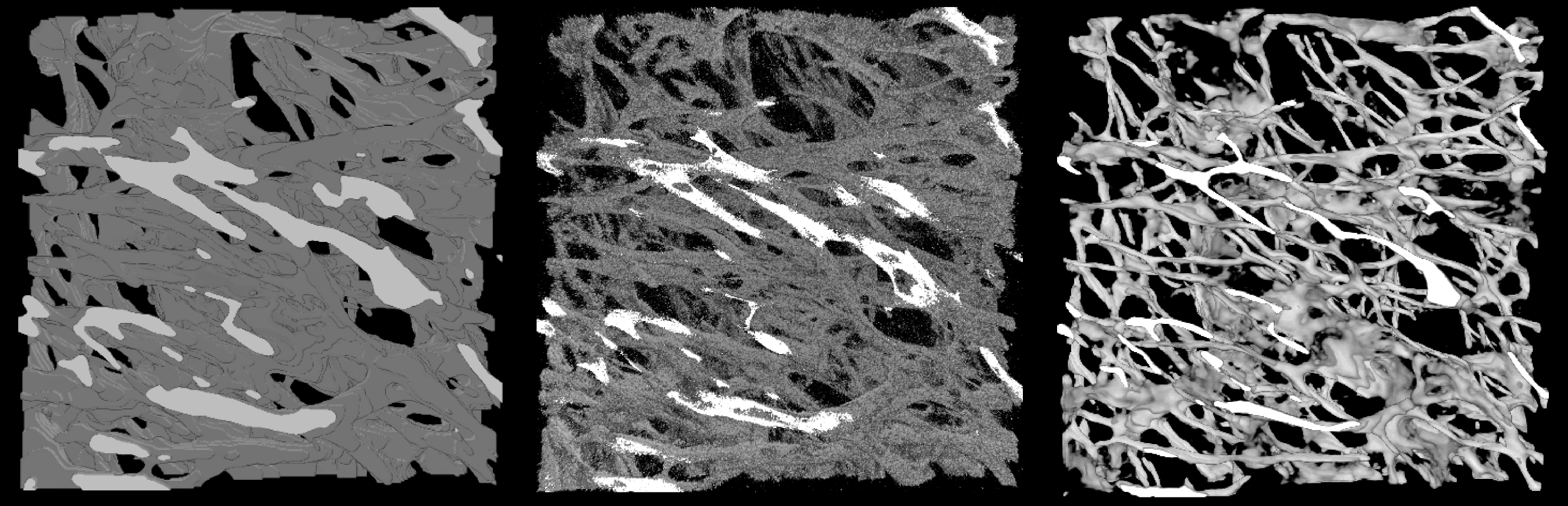}
\caption{Network trained to reproduce segmentation (left), input endothelium channel (center) and network trained to reproduce skeleton (right).}
\label{fig:seg_skel_compare}
\end{figure}

We also test the effectiveness of these architectures on five extremely challenging real image volumes, where tightly packed vasculature makes delineation via segmentation and thinning impossible. We again compare the the Hausdorff average distance between skeletons. These images were taken from a different tumor to those used to construct the training set. All architectures were trained using the weighted binary cross-entropy loss function described in Equation \ref{eq:wbce}. These networks are all trained against the ground truth 3D skeletons. A qualitative demonstration of the benefits of training against the skeleton rather than the segmentation can be seen in Figure \ref{fig:seg_skel_compare}. We see that if the network is trained on the segmentation alone, as in the left pane, many distinct vessels are combined into a single region of segmented vasculature. When we train against the skeletons themselves, as in the right pane, considerably more detail is visible. 

\begin{table}[h]
\title{Skeleton Distance Comparison}
\centering
\begin{tabular}{l|c|c|c}
Architecture & Skeleton Error ($\upmu$m) & Node Distance & Coverage\\ 
\hline
U-2D & 6.53 & 10.48 & 0.98 \\
U-2D+CLSTM (D) & 5.87 & \textbf{9.77} & 0.98 \\
U-2D+CLSTM (S) & \textbf{5.31} & 9.87 & \textbf{0.99} \\
U-3D &  7.53 & 10.68 &  0.92 \\
\end{tabular}
\caption{Results for Experiment 1}
\end{table}

\section{Discussion}

In this paper we have shown that by combining CNN networks with a Convolutional LSTM network we are able to extract truly 3D vascular structures from complex microscopy images while avoiding the complexity of having to perform convolutions in 3D. We have also shown that using skeletal centre-lines as training targets rather than segmentations helps to distinguish between tightly packed structures. In addition to showing this to be an effective method on our task of extracting vessel skeletons, we have explored the role played by various loss functions tailored to this task. In our experiments we found that although the 3D CNN achieves comparable results in terms of segmentation overlap, the U-Net + ConvLSTM architecture shows a clear improvement in terms of ability to recreate skeletal representations of the vasculature. A qualitative comparison of the difference in results between the U-Net2D and the U-Net2D + ConvLSTM architectures can be seen in Figure \ref{fig:results_compare}. In the results from the 2D network we notice that a large number of spurious connections have formed between non-connected parts of the vasculature. We interpret this as being due to the networks inability to separate structures in 3-dimensions. The 2D network performs adequately in regions of low density (for example along the top edge of the image) however begins to fail in the regions of high density in the center. However, our approach performs well in all of these regions as we are able to localize and separate responses in 3D. 

In our comparison of the loss functions, given in Table \ref{tab:loss_func}, we observe that the use of a weighted binary cross-entropy loss produced the best results. While the Dice derived loss function performed well, we found that it was not capable of learning a meaningful model for these extremely sparse skeleton structures. 

In addition to these results it is useful to examine to relative model sizes of the architectures being compared. In the case of the synthetic Experiment 2, we see that the U-Net2D was completely incapable of learning this 3D representation, this is to be expected as contextual 3D information is required. However, the addition of a single ConvLSTM layer, resulting in an increase of about 30\% in the number of trainable weights makes this representation possible. In contrast, the natural extension of the U-Net2D to the U-Net3D requires at least an approximate tripling in the number of parameters, as $3\times3$ kernels are extended to $3\times3\times3$ volumetric kernels. Hybrid convolutional/recurrent architectures of this kind represent a sort of hybrid between 2D and 3D image processing. This is fitting for data of this kind as it also represents a hybrid between 2D and 3D imaging, where images acquired predominantly in 2D are concatenated to form image volumes. Interestingly we noted a slight improvement in performance of the shallow variant of the U-Net + ConvLSTM architecture. We believe this is most likely due to the small training set size, but we believe that this is an important observation as networks that are able to perform with smaller training sets are of great interest to researchers in biomedical applications, where large public datasets will not usually exist for bespoke applications. The results presented here have been achieved using a training set that may reasonably be acquired by a single researcher. 

We believe that the pipeline we have demonstrated here represents a principled approach to information sharing that targets the majority of the computational power of the network to the primary source of information (convolutions in-plane) and forces efficiency in the use of trainable parameters. We envision this approach also being useful in other anisotropic imaging modalities where 3D structures are extracted from a concatenation of higher resolution in-plane images, e.g. cardiac MRI, lung CT or other microscopy applications.  

\section{Acknowledgements}

The research leading to these results has received funding
from the People Programme (Marie Curie Actions) of the
European Unions Seventh Framework Programme (FP7/2007-
2013) under REA grant agreement No 625631. This work
was also supported by Cancer Research UK (CR-UK) grant
numbers C5255/A18085 and C5255/A15935, through the
CRUK Oxford Centre and by CRUK/EPSRC Oxford Cancer Imaging Centre (grant number C5255/A16466). RB acknowledges funding from the
EPSRC Systems Biology Doctoral Training Centre, Oxford
(EP/G03706X/1).

\bibliographystyle{ieeetr}
\bibliography{Mendeley}
\end{document}